\documentclass[sigconf]{acmart}
\AtBeginDocument{%
  }

\copyrightyear{2026}
\acmYear{2026}
\setcopyright{cc}
\setcctype{by}
\acmConference[MM '26]{Proceedings of the 34th ACM International Conference on Multimedia}{November 10--14, 2026}{Rio de Janeiro, Brazil}
\acmBooktitle{Proceedings of the 34th ACM International Conference on Multimedia (MM '26), November 10--14, 2026, Rio de Janeiro, Brazil}
\acmDOI{10.1145/3767308.3836043}
\acmISBN{979-8-4007-2213-4/2026/11}

\usepackage{multirow}
\usepackage{makecell}
\usepackage{subfigure}
\usepackage[table]{xcolor}  % This enables \rowcolor
\usepackage{listings}
\usepackage{pifont}
\usepackage{balance}  % balance the two columns on the last page (ACM CR requirement)
\definecolor{skyblue}{RGB}{135,206,235}

\begin{document}

\title{CaM-Wolf: Causal-Aware Multimodal Agents for Social Deduction Games}

\author{Zheng Zhang}
\orcid{0009-0001-6227-1987}
\affiliation{%
  \institution{The Hong Kong University of Science and Technology (Guangzhou)}
  \city{Guangzhou}
  \country{China}
}
\email{zzhang302@connect.hkust-gz.edu.cn}

\author{Nanjie Yao}
\orcid{0009-0007-4930-7758}
\affiliation{%
  \institution{The Hong Kong University of Science and Technology (Guangzhou)}
  \city{Guangzhou}
  \country{China}
}
\email{nanjieyao@gmail.com}

\author{Jiarui He}
\orcid{0009-0002-0137-7777}
\affiliation{%
  \institution{The Hong Kong University of Science and Technology (Guangzhou)}
  \city{Guangzhou}
  \country{China}
}
\email{jhe218@connect.hkust-gz.edu.cn}

\author{Deheng Ye}
\orcid{0000-0002-1754-1837}
\affiliation{%
 \institution{Nanyang Technological University}
 \city{Singapore}
 \country{Singapore}
}
\email{775050607@qq.com}

\author{Peilin Zhao}
\orcid{0000-0001-8543-3953}
\affiliation{%
  \institution{Shanghai Jiao Tong University}
  \city{Shanghai}
  \country{China}
}
\email{peilinzhao@hotmail.com}

\author{Hao Wang}
\orcid{0000-0002-3086-3128}
\authornote{Corresponding author.}
\affiliation{%
  \institution{The Hong Kong University of Science and Technology (Guangzhou)}
  \city{Guangzhou}
  \country{China}
}
\email{haowang@hkust-gz.edu.cn}

\renewcommand{\shortauthors}{Zheng Zhang et al.}

\begin{abstract}
Social deduction games (SDGs) such as Werewolf have become challenging testbeds for AI agents. These games require complex social skills such as reasoning, deception, and collaboration. While recent advances in large language models (LLMs) have driven significant progress in SDG agents, current approaches are predominantly text-based, overlooking the multimodal nature that is fundamental to human social interaction. To bridge this gap, we introduce CaM-Wolf, the first SDG agent that integrates multimodal perception and generation. CaM-Wolf processes video inputs from other players, employs a causal-aware Reasoner trained via reinforcement learning to establish logical chains between observable behaviors and hidden roles, and presents itself through an animated avatar. Our experiments and user study show that CaM-Wolf achieves superior agent gameplay performance and enhances the quality of human-AI interaction. This work represents a significant advancement towards creating more human-like AI agents capable of participating in nuanced social dynamics. Our code is available at \url{https://3dagentworld.github.io/avatar_wolf}.
\end{abstract}




\maketitle

\section{Introduction}

\begin{figure}[ht]
    \centering
    \includegraphics[width=\columnwidth]{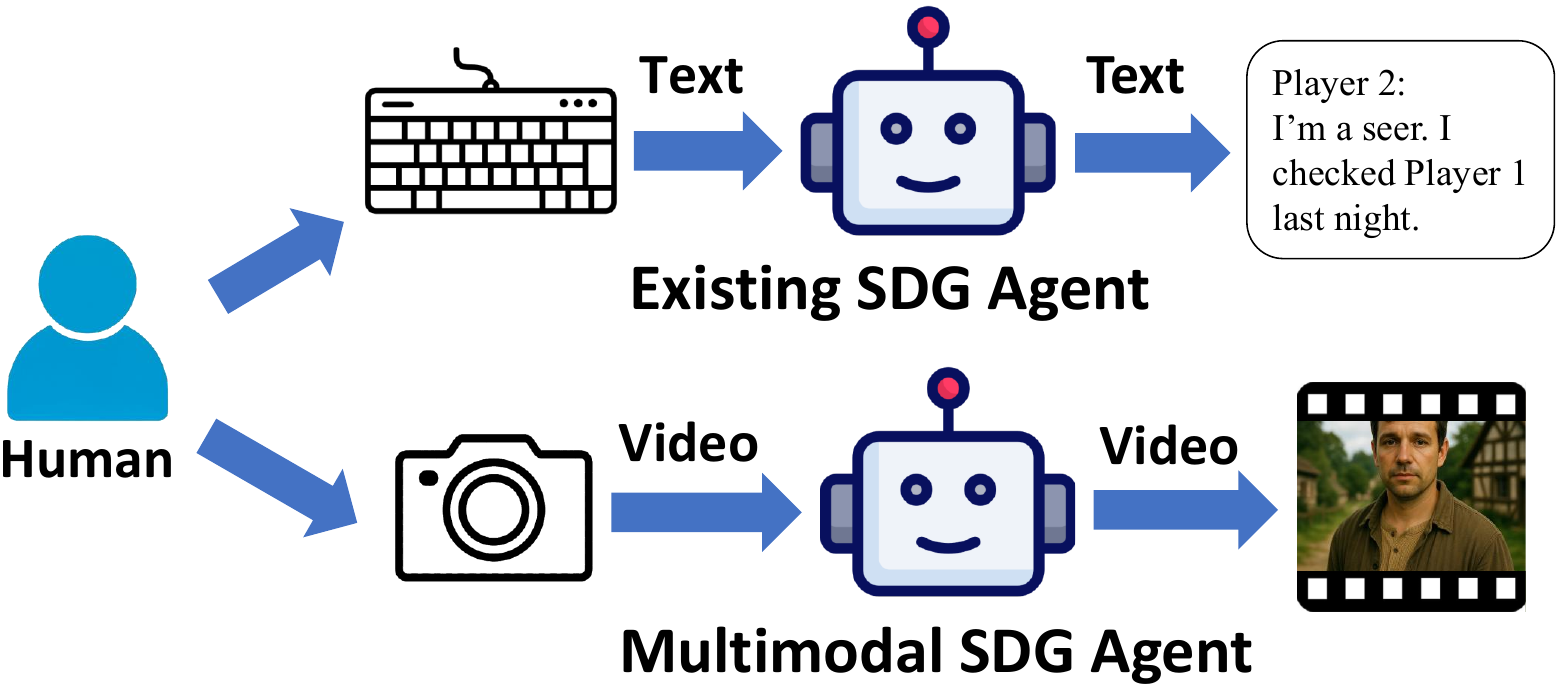}
    % \vspace{-15pt}
    \caption{\textbf{Different interaction paradigms of SDG agents.} When interacting with existing SDG agents, human players must input their statements through keyboard text entry, and agents generate textual responses. In contrast, our proposed multimodal SDG agents capture human players' video input during their speeches and generates videos as responses.}
    \label{fig:motivation}
    % \vspace{-15pt}
\end{figure}

The rapid development of large language models (LLMs) has driven significant progress in creating intelligent agents that can simulate human-like decision-making and social behaviors. These LLM-based agents have shown strong performance across various domains, from controlling computer desktops \citep{nayak2025uivision, wang2025oscar, wang-etal-2025-ponder} and web browsing~\citep{luo-etal-2025-browsing, 10.5555/3692070.3694608} to playing complex video games like StarCraft \citep{ma2024large, shao2024swarmbrainembodiedagentrealtime} and Minecraft \citep{wang2024voyager, fu2025vistawisebuildingcosteffectiveagent, chai2025causalmacecausalityempoweredmultiagents, he2025robust, zhou2026htac, dai2024mamba}. While these tasks demonstrate impressive capabilities, they often rely primarily on observable information and predefined action spaces. To evaluate an agent's ability to handle hidden information and social reasoning, social deduction games (SDGs) have been adopted as challenging testbeds that better reflect the complexity of human social interaction.

In SDGs such as Werewolf \citep{10888525, xu2023exploring, 10.5555/3692070.3694355, wu2024enhancereasoninglargelanguage}, Avalon \citep{light2023from, wang2023avalonsgamethoughtsbattle, shi2023cooperationflyexploringlanguage, lan-etal-2024-llm}, and Jubensha \citep{10.1145/3721121, wu-etal-2024-deciphering}, players are secretly assigned different roles with competing objectives, and have to identify each other's hidden identities through discussion. Unlike many other games where actions are chosen from a limited set of options, SDGs rely heavily on free-form natural language interactions where players persuade, deceive, and collaborate through conversation.

Despite notable progress in developing LLM agents for SDGs, current approaches exhibit significant limitations. As shown in Figure \ref{fig:motivation}, most existing agents \citep{10888525, xu2023exploring, 10.5555/3692070.3694355, wu2024enhancereasoninglargelanguage, jin2024learning, lan-etal-2024-llm, zheng-etal-2026-stackelberg} are constrained to processing purely textual information. When these text-based agents interact with human players, users can only see agent statements displayed as text and must input their own responses through text entry, creating an interaction paradigm that diverges substantially from natural human communication. This text-only interface limits the validity of human-AI interaction studies and fails to capture the rich multimodal information exchange that characterizes authentic social deduction scenarios. While some research has begun incorporating multimodal data in SDG \citep{lai-etal-2023-werewolf, 10657493}, these approaches primarily focus on retrospective analysis rather than active gameplay participation.

Additionally, existing methods typically concentrate on communication strategy selection, overlooking the crucial aspect of reasoning about hidden roles, which is also a core challenge in SDGs. Also, current LLM agents often make inferences based on internal biases \citep{10.1145/3703155, xu2025hallucinationinevitableinnatelimitation}, leading to unreliable reasoning that lacks explicit supporting evidence. Given that each player's behaviors in SDGs are inherently shaped by their hidden roles, we propose that causal relations should exist between players' observable behaviors and their concealed identities. By establishing these causal connections, we can build explicit logical chains that trace from specific observed behaviors to role predictions, ensuring that every inference is supported by concrete evidence rather than arbitrary speculation. This assumption enables us to use causal discovery~\citep{10.1145/3595380, wu-etal-2024-decot} to guide more accurate social deduction reasoning.

In this paper, we focus on the Werewolf game. In Werewolf, players are divided into Team Village and Team Werewolf. The game alternates between night phases, where werewolves secretly eliminate a player, and day phases, where all surviving players discuss and vote to eliminate a suspected werewolf. This dynamic creates an ideal testbed for exploring complex social deduction and multimodal interaction.

To this end, we present CaM-Wolf, a novel framework that enhances LLM agents for Werewolf by integrating multimodal perception and generation. CaM-Wolf takes users' video inputs and generates transcriptions of their speech content along with descriptions of their facial expressions and gestures. During output generation, CaM-Wolf presents itself through a digital avatar that combines speech synthesis and video generation, enhancing the naturalness and immersiveness of human-AI interaction.

We further develop a causal-aware Reasoner trained via reinforcement learning. We design a counterfactual premise intervention mechanism to verify the faithfulness of the agent's reasoning, and formulate causal rewards to optimize the model using Group Relative Policy Optimization (GRPO) \citep{shao2024deepseekmathpushinglimitsmathematical}. This training paradigm enables CaM-Wolf to intrinsically generate faithful and evidence-grounded reasoning, making more informed decisions in subsequent communication strategy selection and response formulation, leading to more contextually-aware gameplay.

Our main contributions can be summarized as follows:

\begin{itemize}
    \item To the best of our knowledge, we are the first to develop a framework that integrates multimodal perception and generation into SDG agents, enabling more natural and immersive human-AI interaction.
    \item We propose a causal-aware Reasoner trained via reinforcement learning with causal intervention rewards, enabling the agent to intrinsically generate faithful and evidence-grounded reasoning about other players' true identities.
    \item We demonstrate the effectiveness of our approach through comprehensive evaluation in both agent-versus-agent simulations and studies with human players, showing improvements in gameplay performance and human-AI interaction.
\end{itemize}

\begin{figure*}
    \centering
    \includegraphics[width=\textwidth]{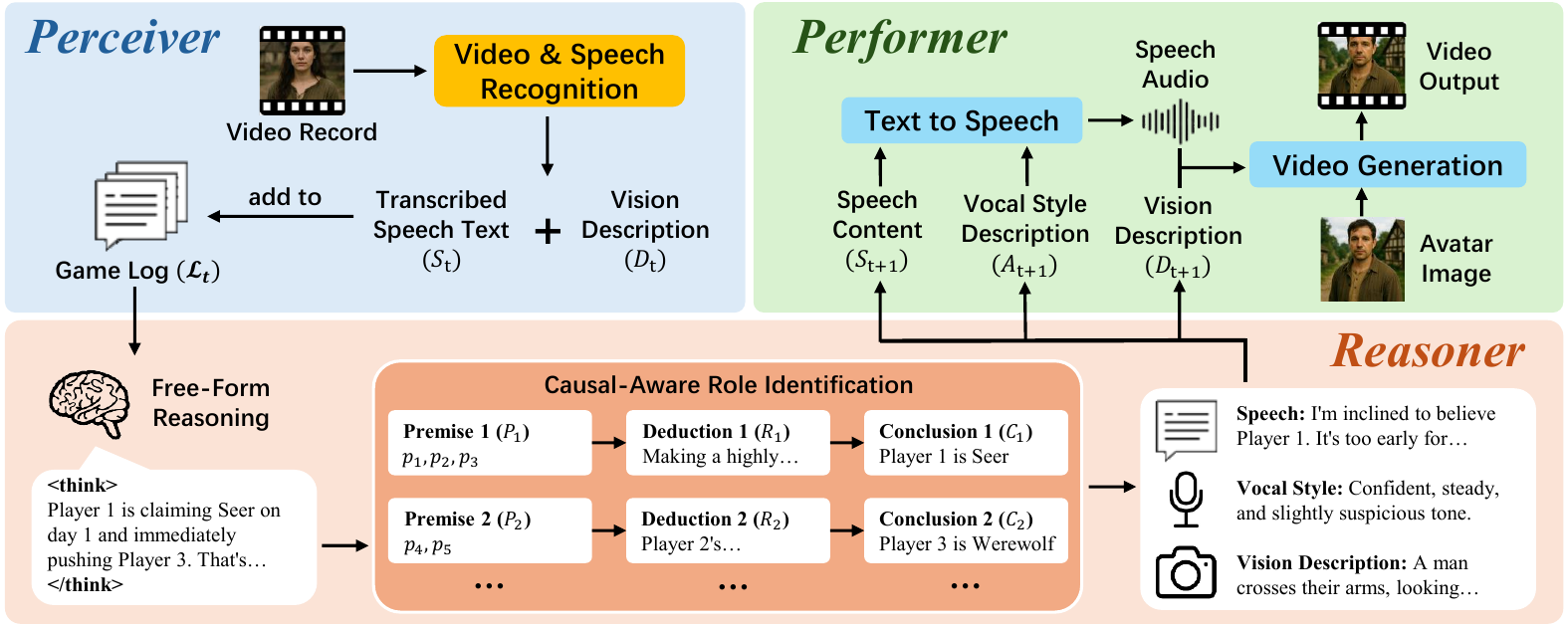}
    % \vspace{-12pt}
    \caption{\textbf{The overall framework of CaM-Wolf.} The Perceiver processes video inputs from human players, extracting transcribed speech text and vision description. The Reasoner conducts causal-aware role identification and generates responses. The Performer presents the responses through an animated avatar.}
    \label{fig:framework}
    % \vspace{-6pt}
\end{figure*}

\section{Related Work}

\subsection{Social Deduction Game Agents}

The evolution of AI agents for social deduction games (SDGs) has progressed through two phases. Initial investigations into SDG agents \citep{Hirata2016WerewolfGM, Nakamura2016ConstructingAH, Wang2018ApplicationOD, 10.5555/3454287.3454400} rely on deterministic rule-based frameworks or traditional machine learning techniques for decision-making processes. These frameworks operate through rigid communication protocols rather than flexible natural language interaction, limiting their capacity for nuanced social reasoning.

The advent of LLMs has facilitated the development of more sophisticated agents across diverse SDGs. For Avalon, \citet{lan-etal-2024-llm} leverage manually crafted system prompts to direct LLM behavior during gameplay, and DeepRole \citep{10.5555/3454287.3454400} integrates counterfactual regret minimization algorithms with neural value networks optimized through iterative self-play training. For traditional Werewolf, \citet{xu2023exploring} generate multiple strategic action candidates via deductive reasoning and employ reinforcement learning to enhance action selection. Building upon this, \citet{wu2024enhancereasoninglargelanguage} incorporate a Thinker module to augment reasoning capabilities for complex logical analysis. For One Night Ultimate Werewolf (ONUW), \citet{jin2024learning} employ reinforcement learning to train agents in selecting from predefined communication strategies.

However, existing SDG agents merely depend on textual information. This constraint represents a significant departure from authentic human social interaction. The absence of multimodal perception and generation capabilities in current agents creates a substantial gap between artificial and human gameplay experiences, limiting their applicability in realistic scenarios.

\subsection{Multimodal Social Interaction}

The intersection of multimodal analysis and social dynamics has garnered increasing research attention. For example, Ego4D \citep{Grauman_2022_CVPR} provides a framework for analyzing social attention patterns through integrated video and audio modalities. In SDG contexts, \citet{10657493} explore modeling challenges by developing densely aligned language-visual representations to capture fine-grained behavioral dynamics, while \citet{lai-etal-2023-werewolf} conduct comprehensive analysis of ONUW gameplay recordings, identifying correlations between non-verbal behavioral patterns and deceptive strategies.

These works primarily focus on retrospective analysis of social interactions rather than active participation in gameplay. They typically examine recorded gameplay logs to extract behavioral patterns or identify deception indicators, without leveraging these insights for decision-making during active gameplay scenarios. In contrast, our work represents a fundamental departure from this paradigm by integrating both multimodal perception and generation into the agent's pipeline, enabling more authentic gameplay experiences.

\subsection{Causal Discovery}

Causal discovery can be broadly categorized into two paradigms: conventional statistical approaches and LLM-based techniques. Traditional statistical methods primarily employ algorithmic and computational frameworks to identify causal relationships within data structures. Many approaches \citep{chickering2015selective, ramsey2015scaling, xiang2013lasso} evaluate scoring metrics across potential edge configurations, seeking optimal graph structures that maximize data fitting. Besides, some methods \citep{spirtes2001causation, spirtes2013causal} utilize conditional independence testing to detect causal architectures by examining variable interdependencies.

The emergence of LLMs has introduced novel possibilities for causal discovery applications. \citet{cohrs2024large} explore LLMs as substitutes for traditional conditional independence testing within established causal discovery pipelines, and \citet{long2023causal} focus on constructing causal graphs that transcend Markov equivalence class limitations. Furthermore, data-free methodologies~\citep{kiciman2023causal, zevceviccausal, zhang2024causal} demonstrate that LLMs can extract causal structures through metadata interpretation and natural language processing, mirroring expert domain knowledge application in causal model construction.

The inherent structure of SDGs creates natural causal relationships between players' hidden roles and their observable behaviors, as each role constrains and influences player actions in predictable ways. By integrating causal discovery into a reinforcement learning framework, our agent can intrinsically learn to identify these causal connections and develop more systematic approaches to social deduction, moving to a deeper understanding of the underlying game dynamics.

\section{Method}

\subsection{Overview}

CaM-Wolf consists of three modules that enable multimodal social deduction gameplay. As illustrated in Figure~\ref{fig:framework}, the framework processes multimodal inputs and generates avatar-based responses.

The \textbf{Perceiver} converts video inputs from human players into textual descriptions, extracting both speech content and visual cues such as facial expressions and gestures.

The \textbf{Reasoner} then conducts causal-aware role identification by systematically analyzing the relationship between premises to conclusions. It infers player roles and generates the appropriate speech content, vocal style, and vision description.

The \textbf{Performer} takes these generated descriptions and presents the response through an animated avatar that combines speech synthesis and video generation.

\begin{figure*}
    \centering
    \includegraphics[width=\textwidth]{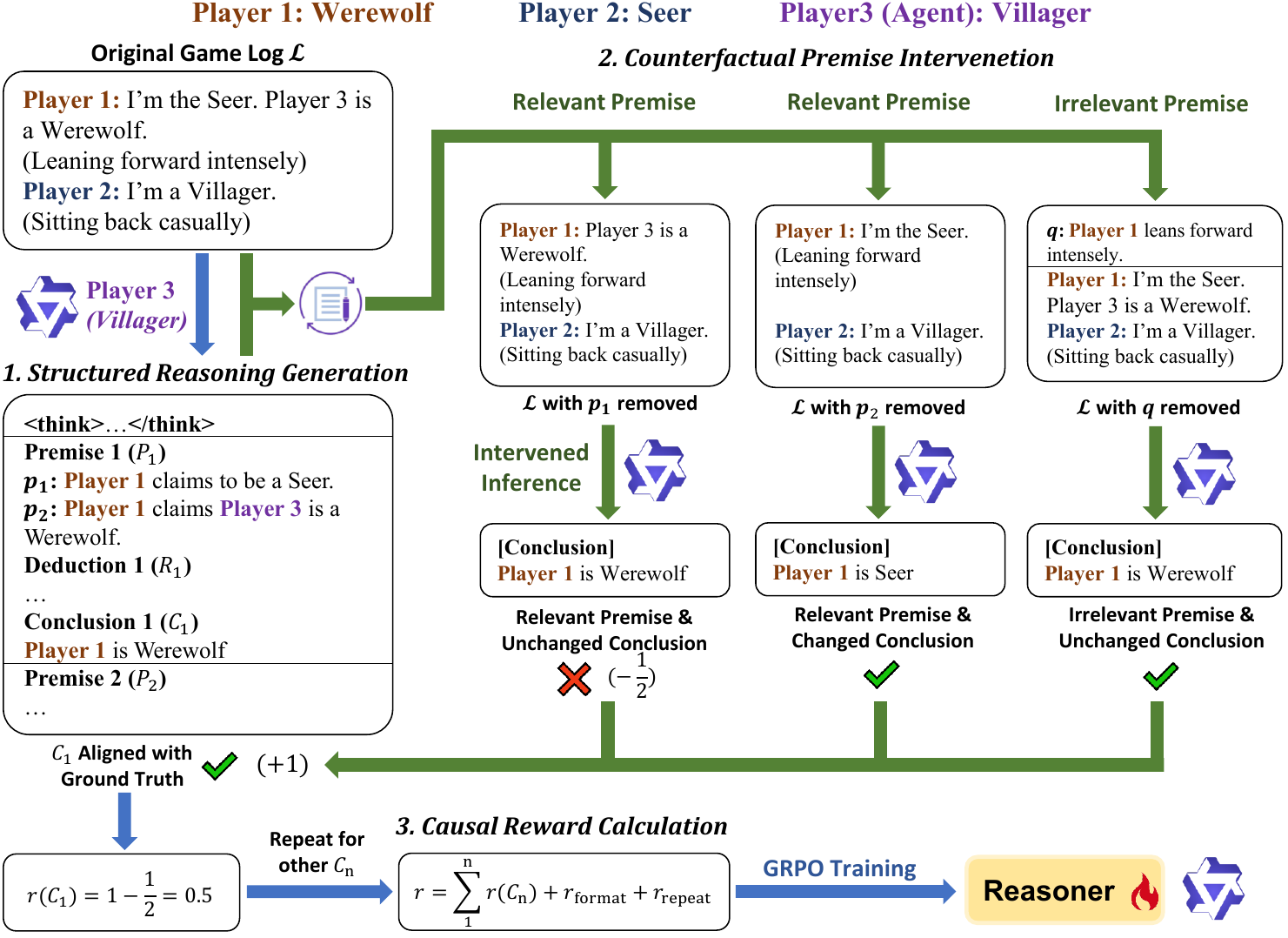}
    % \vspace{-8pt}
    \caption{\textbf{Overview of the Reasoner's training process.} The Reasoner first generates structured reasoning with explicit premise-deduction-conclusion chains. Then, counterfactual premise intervention is performed by removing relevant and irrelevant premises from the game log respectively and re-inferring the same player's role. If the conclusion changes, the removed premise is confirmed as genuine causal evidence; otherwise, it is considered a spurious premise. Finally, causal rewards are calculated based on the intervention results to optimize the model.}
    \label{fig:causal_example}
    % \vspace{-10pt}
\end{figure*}

\subsection{Perceiver}

The Perceiver processes multimodal video inputs from human players to extract comprehensive information including speech content and description of other behaviors. It transforms raw video data into structured textual representations for subsequent reasoning.

Given a video input $V_t$ from a player at time $t$, we employ Qwen2.5-Omni-7B\footnote{\url{https://huggingface.co/Qwen/Qwen2.5-Omni-7B}} \citep{xu2025qwen25omnitechnicalreport} to transcribe audio content from the video input, as well as analyze visual components and generate textual descriptions of the player's gestures and facial expressions:
\begin{equation}
S_t, D_t = f_{\text{qwen}}(V_t),
\end{equation}
where $V_t$ represents the video, $S_t$ is the transcribed speech text, and $D_t$ represents the generated vision description.

The complete multimodal information extracted at time $t$ is then formalized as:
\begin{equation}
\label{eq:merge}
I_t = \{S_t, D_t\},
\end{equation}
where $I_t$ encompasses both linguistic and visual information.

These extracted information components are subsequently integrated into the game log $\mathcal{L}$ as structured entries, ensuring temporal ordering and player attribution for subsequent causal analysis. At each time step $t$, the game log is updated as:

\begin{equation}
\mathcal{L}_{t} = \mathcal{L}_{t-1} \cup I_t,
\end{equation}
where $\mathcal{L}_{t}$ represents the updated game log at time $t$, maintaining a complete temporal record of all interactions.

\subsection{Reasoner}
\label{sec:reasoner}

Our Reasoner is a causal-aware reasoning LLM trained via reinforcement learning to intrinsically generate faithful and evidence-grounded reasoning. We optimize the reasoning process using Group Relative Policy Optimization (GRPO)~\citep{shao2024deepseekmathpushinglimitsmathematical} guided by causal intervention rewards. As illustrated in Figure \ref{fig:causal_example}, the training process involves three steps: (1) generating structured reasoning with explicit premise-deduction-conclusion chains, (2) performing counterfactual premise intervention to verify reasoning faithfulness, and (3) calculating causal rewards for model optimization.

During inference, the trained LLM only needs to execute the first step, and the generated responses are directly passed to the Performer.

\begin{figure}
    \centering
    \includegraphics[width=\columnwidth]{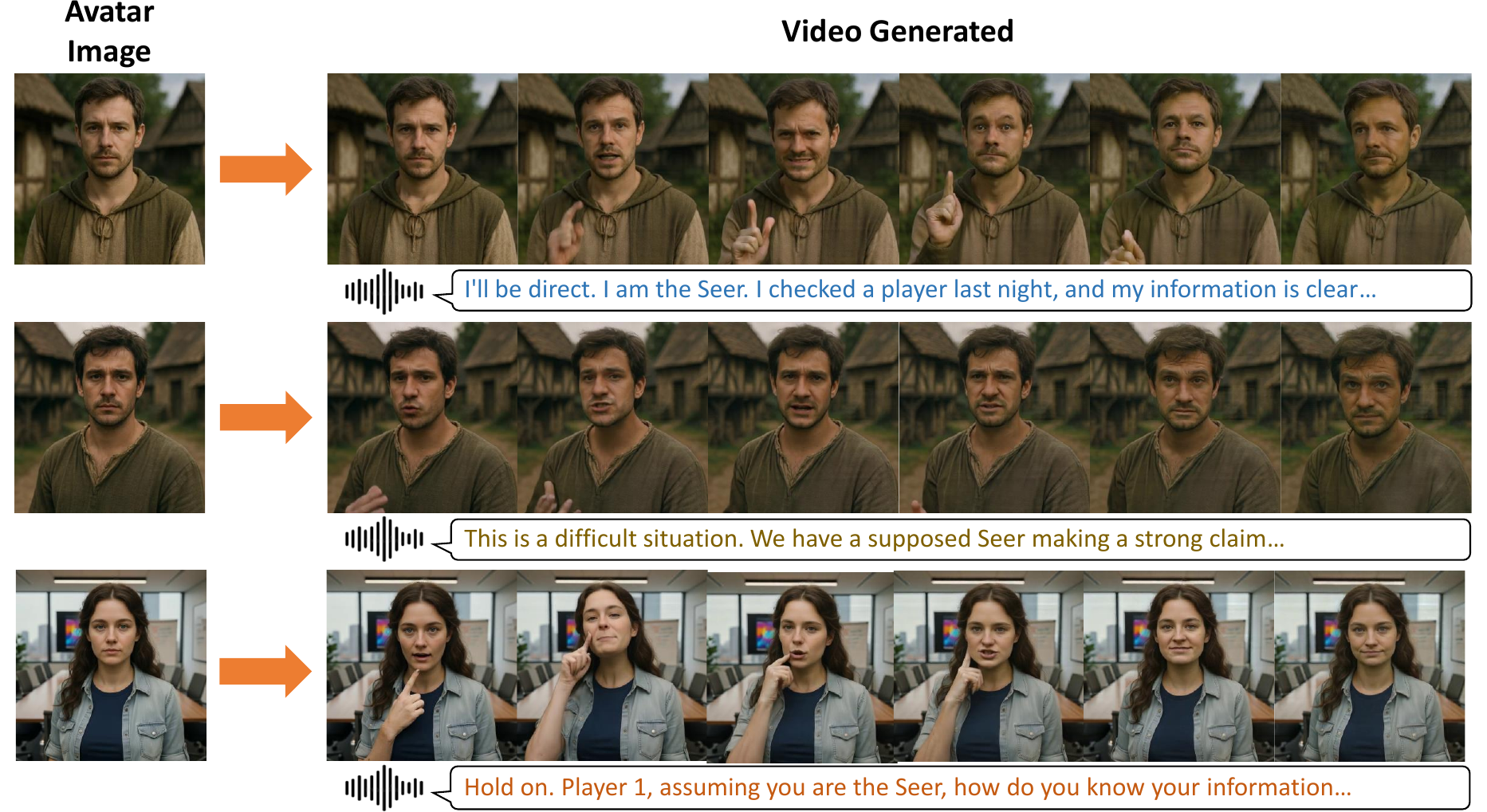}
    \vspace{-8pt}
    \caption{\textbf{Examples of generated videos.} The left side shows avatar images generated by GPT-4o-Image for each agent player before the game starts. The right side presents video examples of these avatars speaking during gameplay.}
    \label{fig:visualize}
    \vspace{-8pt}
\end{figure}

\subsubsection{Structured Reasoning Generation.}
\label{sec:reasoning_generation}
To enable precise reward assignment and evaluation, we constrain the agent's reasoning output to a structured format. As shown in Figure \ref{fig:framework}, during each speaking turn, given the game rules $\mathcal{R}$ and current game log $\mathcal{L}_t$, the agent first performs free-form reasoning within a \texttt{<think>} block. Subsequently, it outputs a sequence of $n$ role identifications $\{d_1, d_2, \ldots, d_n\}$, where each role identification $d_i$ is a triple:
\begin{equation}
\label{eq:deduction}
d_i = (P_i, R_i, C_i),
\end{equation}
where $P_i = \{p_1, p_2, \ldots, p_{|P_i|}\}$ is a set of premises citing specific behavioral observations (including multimodal descriptions) from $\mathcal{L}_t$, $R_i$ is the deduction process, and $C_i$ is a conclusion about the role of another player. Finally, the agent produces its speech content $S_{t+1}$, a vocal style description $A_{t+1}$, and a vision description $D_{t+1}$ for avatar generation.

\subsubsection{Counterfactual Premise Intervention.}
\label{sec:premise_intervention}
To evaluate the faithfulness of the generated role identifications, we introduce an automated counterfactual intervention mechanism. For each correctly identified role $d_i$ with conclusion $C_i$ (where $C_i$ aligns with the ground truth), we perform two types of interventions on the game log $\mathcal{L}_t$ using a lightweight LLM (Qwen2.5-14B-Instruct):

\noindent\textbf{Relevant premise intervention.} For each cited premise $p_j \in P_i$ of conclusion $C_i$, we remove the corresponding behavioral observation in the game log:
\begin{equation}
\label{eq:relevant_intervene}
\mathcal{L}_t^{-p_j} = f_{\text{intervene}}(p_j, \mathcal{L}_t).
\end{equation}

\noindent\textbf{Irrelevant premise intervention.} We randomly select an observation $q \notin P_i$ from the premises of other role identifications and remove it:
\begin{equation}
\label{eq:irrelevant_intervene}
\mathcal{L}_t^{-q} = f_{\text{intervene}}(q, \mathcal{L}_t).
\end{equation}

The LLM then re-infers on each intervened game log to produce a new role prediction $C_i'$ for the same player. We compare $C_i'$ with the original prediction $C_i$ to determine whether the intervention has changed the conclusion ($\delta=1$) or not ($\delta=0$).

\subsubsection{Causal Reward Calculation.}
Based on the intervention results, we design a composite reward function $r$ to optimize the reasoning process. Let $y$ denote the ground-truth role assignments.

\noindent\textbf{Correctness reward} $r_{\text{correct}}(C_i)$: A reward of $+1$ is assigned if the predicted conclusion $C_i$ aligns with the ground-truth role assignment $y$; otherwise, a penalty of $-1$ is applied.

\noindent\textbf{Faithfulness reward} $r_{\text{faith}}(C_i)$: For each correctly predicted $C_i$, we evaluate two aspects. First, if removing a relevant premise $p_j$ fails to change the conclusion ($\delta = 0$), a penalty of $-\frac{1}{|P_i|}$ is applied per such premise, discouraging the model from citing irrelevant evidence. Second, if removing an irrelevant premise causes the conclusion to change ($\delta = 1$), a penalty of $-0.5$ is applied, encouraging the model to explicitly cite all premises it relies on.

\noindent\textbf{Format reward} $r_{\text{format}}$: A penalty of $-1$ is applied when the output violates the predefined structured template.

\noindent\textbf{Repetition penalty} $r_{\text{repeat}}$: For each of the other players, the role identification is permitted to appear at most once. A penalty of $-1$ is applied for each redundant occurrence.

The total reward for a single response is:
\begin{equation}
\label{eq:reward}
r = \sum_{i=1}^{n} \left( r_{\text{correct}}(C_i) + r_{\text{faith}}(C_i) \right) + r_{\text{format}} + r_{\text{repeat}}.
\end{equation}

\begin{table*}[ht]
\caption{\textbf{Performance comparison on the Werewolf game.} We evaluate each baseline against ReAct under different team assignments, conducting 50 games for each setting. In each game, players from the same team use the same agent type. For the training-based method (LSPO), we reproduce it on Qwen2.5-72B-Instruct using an identical amount of training data.}
\vspace{-8pt}
\label{tab:main_results}
\centering
\begin{tabular}{llccc}
\toprule
\multirow{2}{*}{\textbf{Method}} & \multirow{2}{*}{\textbf{Opponent}} & \multicolumn{3}{c}{\textbf{Win Rate (\%)}} \\
\cmidrule(lr){3-5}
& & \textbf{Team Villager} & \textbf{Team Werewolf} & \textbf{Overall} \\
\midrule
ReAct (Qwen2.5-72B-Instruct) & ReAct (GPT-4o-mini) & 36.0 & 88.0 & 62.0 \\
ReAct (Qwen2.5-72B-Instruct) & ReAct (GPT-4o) & 18.0 & 70.0 & 44.0 \\
\midrule
ReAct (GPT-4o) & ReAct (GPT-4o) & 28.0 & 72.0 & 50.0 \\
ReCon (GPT-4o) & ReAct (GPT-4o) & 34.0 & 80.0 & 57.0 \\
SLA (GPT-4o) & ReAct (GPT-4o) & 34.0 & \textbf{86.0} & \textbf{60.0} \\
LSPO (Qwen2.5-72B-Instruct) & ReAct (GPT-4o) & 24.0 & 72.0 & 48.0 \\
CaM-Wolf (Ours) & ReAct (GPT-4o) & \textbf{38.0} & 82.0 & \textbf{60.0} \\
\midrule
ReAct (Qwen2.5-72B-Instruct) & ReAct (Qwen2.5-72B-Instruct) & 24.0 & 76.0 & 50.0 \\
ReCon (Qwen2.5-72B-Instruct) & ReAct (Qwen2.5-72B-Instruct) & 28.0 & 80.0 & 54.0 \\
SLA (Qwen2.5-72B-Instruct) & ReAct (Qwen2.5-72B-Instruct) & 30.0 & 84.0 & 57.0 \\
LSPO (Qwen2.5-72B-Instruct) & ReAct (Qwen2.5-72B-Instruct) & 34.0 & 82.0 & 58.0 \\
CaM-Wolf (Ours) & ReAct (Qwen2.5-72B-Instruct) & \textbf{52.0} & \textbf{88.0} & \textbf{70.0} \\
\bottomrule
\end{tabular}
\vspace{-8pt}
\end{table*}

\subsubsection{Model Optimization}

We optimize the Reasoner $\pi_\theta$ using Group Relative Policy Optimization (GRPO) \citep{shao2024deepseekmathpushinglimitsmathematical}, which does not require a separate critic model. For each speaking turn, we sample a group of $G$ candidate responses $\{o_1, \ldots, o_G\}$ from $\pi_\theta$ and compute their rewards $\{r_1, \ldots, r_G\}$ using Equation~\eqref{eq:reward}.

We first calculate normalized advantages:
\begin{equation}
\hat{A}_g = \frac{r_g - \mu_G}{\sigma_G},
\end{equation}
where $\mu_G$ and $\sigma_G$ are the mean and standard deviation of rewards $\{r_1, \ldots, r_G\}$.

We then update the policy by maximizing the GRPO objective:
\begin{equation}
\label{eq:grpo}
\mathcal{J}_{\text{GRPO}}(\theta) = \mathbb{E} \left[ \frac{1}{G} \sum_{g=1}^{G} \mathcal{L}_g - \beta D_{\text{KL}}(\pi_\theta \| \pi_{\text{ref}}) \right],
\end{equation}
where $\mathcal{L}_g$ is the clipped surrogate objective:
\begin{equation}
\mathcal{L}_g = \min\!\left( \rho_g \hat{A}_g,\; \text{clip}(\rho_g, 1{-}\epsilon, 1{+}\epsilon)\, \hat{A}_g \right),
\end{equation}
and the importance ratio $\rho_g$ is defined as:
\begin{equation}
\rho_g = \frac{\pi_\theta(o_g \mid x)}{\pi_{\theta_{\text{old}}}(o_g \mid x)},
\end{equation}
where $x$ denotes the input context for the current speaking turn. Here, $\epsilon$ controls the clipping range for stable updates, and $\beta$ weights the KL divergence regularization against a reference model $\pi_{\text{ref}}$ to prevent excessive policy deviation.

Through this training process, the Reasoner iteratively improves both the accuracy and the logical faithfulness of its reasoning, learning to conduct causal-aware role identification within a single response.

\subsection{Performer}

The Performer generates the agent's response (including the speech content $S_{t+1}$, vocal style description $A_{t+1}$, and vision description $D_{t+1}$) and presents it through a multimodal avatar, combining speech synthesis and visual animation.

\noindent\textbf{Audio Synthesis.} The textual speech content $S_{t+1}$ and vocal descriptions $A_{t+1}$ are processed through EmotiVoice\footnote{\url{https://github.com/netease-youdao/EmotiVoice}} to generate speech audio:
\begin{equation}
\label{eq:audio_gen}
V_{t+1}^{\text{audio}} = f_{\text{emotivoice}}(S_{t+1}, A_{t+1}).
\end{equation}

\noindent\textbf{Video Generation.} The synthesized audio $V_{t+1}^{\text{audio}}$, a predefined avatar image $F_{\text{avatar}}$ (generated using GPT-4o-Image before the game starts), and the vision description $D_{t+1}$ are used as inputs for OmniAvatar \citep{gan2025omniavatarefficientaudiodrivenavatar} to produce the final avatar video:
\begin{equation}
V_{t+1} = f_{\text{omniavatar}}(F_{\text{avatar}}, V_{t+1}^{\text{audio}}, D_{t+1}),
\end{equation}
where $V_{t+1}$ represents the complete multimodal avatar presentation that synchronizes lip movements, facial expressions, and gestures with the spoken content.

This integrated approach enables CaM-Wolf to deliver natural and engaging multimodal interactions that more closely approximate human-like communication in the gameplay. Examples of the generated avatar videos are shown in Figure~\ref{fig:visualize}.

\section{Experiments}

\subsection{Implementation Details}
\label{sec:ep_setup}

We evaluate our approach on the Werewolf game. In each game, there are two werewolves, one seer, one guardian, and three villagers. The werewolf belongs to Team Werewolf, while all other roles belong to Team Village.

The gameplay alternates between night and day phases. In the night phase, werewolves secretly eliminate a target, while the seer and the guardian use their unique abilities to inspect or protect players. In the day phase, all surviving players engage in sequential discussions and cast votes to eliminate a suspected player. Team Village wins by correctly identifying and eliminating all werewolves through majority voting. Team Werewolf wins by surviving until they equal or outnumber the villagers. Detailed game rules are provided in the supplementary materials.

\subsubsection{Training Scheme}
\label{sec:training_scheme}

We employ Qwen2.5-72B-Instruct\footnote{\url{https://huggingface.co/Qwen/Qwen2.5-72B-Instruct}} \citep{qwen2025qwen25technicalreport} as the base model for the Reasoner and optimize it using the GRPO reinforcement learning framework described in Section~\ref{sec:reasoner}. To construct the training dataset, we first generate 500 game logs via self-play using the vanilla agent framework, where each agent is powered by the Qwen2.5-72B-Instruct model. From the resulting game logs, we randomly sample 3,000 speaking turns to serve as our training data.

Our training pipeline is implemented based on the \texttt{verl} framework\footnote{\url{https://github.com/verl-project/verl}} \citep{10.1145/3689031.3696075}. For the GRPO hyperparameter settings, we configure the sample size to $G=8$, the clipping parameter to $\epsilon=0.2$, and the KL divergence coefficient to $\beta=0.04$. The total batch size per step is set to 128. To alleviate the computational burden, we adopt LoRA \citep{hu2022lora} with a rank of 16 during training.

To accelerate the reward calculation process, we utilize the lightweight Qwen2.5-14B-Instruct\footnote{\url{https://huggingface.co/Qwen/Qwen2.5-14B-Instruct}} model to execute the counterfactual premise intervention detailed in Section~\ref{sec:premise_intervention}. Specifically, we prompt the model to output the modifications in a \texttt{diff} format, which significantly reduces the generation length. Furthermore, as illustrated in Figure~\ref{fig:causal_example}, because the premise intervention and subsequent re-inference for each individual premise are strictly independent, we fully parallelize this process to enhance computational efficiency. 

The training is conducted with a learning rate of $1 \times 10^{-6}$ across 16 H20 GPUs. The entire optimization process spans 2 epochs and requires approximately 6 hours to complete.

\subsubsection{Evaluation Setup}

We evaluate our approach against several baseline agents on the Werewolf game, including ReAct \citep{yao2023react}, ReCon~\citep{wang-etal-2024-boosting-llm}, SLA \citep{10.5555/3692070.3694355}, and LSPO \citep{xu2025learning}. During evaluation, unless otherwise specified, we use GPT-4o as the backend LLM for prompt-based methods (ReAct, ReCon, and SLA). For the training-based method (LSPO), we reproduce it on Qwen2.5-72B-Instruct using an identical amount of training data.

In agent-versus-agent experiments, to ensure fair comparison, we directly provide other baseline agents with the speech content $S_{t+1}$ and vision description $D_{t+1}$ generated by our agent in Section~\ref{sec:reasoning_generation} as inputs, bypassing the multimodal processing procedure to conduct pure textual gameplay.

In human-agent mixed gameplay, we integrate our Perceiver and Performer into other baseline agents to enable complete multimodal interaction capabilities. To facilitate this multimodal setting on the human side, each human participant is seated individually at a computer where their video and audio are recorded.

\subsection{Main Results}

\begin{table}[t]
\caption{\textbf{Role identification accuracy.} Using the game logs from the games against ReAct (Qwen2.5-72B-Instruct) in Table~\ref{tab:main_results}, we explicitly prompt each agent to predict the roles of all other players at the end of the second day-phase discussion and report the percentage of correct predictions.}
\vspace{-8pt}
\label{tab:role_identification}
\centering
\begin{tabular}{lc}
\toprule
\textbf{Method} & \textbf{Accuracy (\%)} \\
\midrule
ReAct (Qwen2.5-72B-Instruct) & 22.4 \\
ReCon (Qwen2.5-72B-Instruct) & 25.8 \\
SLA (Qwen2.5-72B-Instruct) & 27.2 \\
LSPO (Qwen2.5-72B-Instruct) & 27.6 \\
CaM-Wolf (Ours) & \textbf{32.6} \\
\bottomrule
\end{tabular}
\vspace{-8pt}
\end{table}

\begin{table}[t]
\caption{\textbf{Ablation study.} Each variant plays against ReAct (GPT-4o) as either Team Village or Team Werewolf. We conduct 50 games for each setting to evaluate the contribution of each individual component.}
\vspace{-8pt}
\label{tab:ablation}
\centering
\begin{tabular}{lccc}
\toprule
\multirow{2}{*}{\textbf{Variant}} & \multicolumn{3}{c}{\textbf{Win Rate (\%)}} \\
\cmidrule(lr){2-4}
& \textbf{Village} & \textbf{Werewolf} & \textbf{Overall} \\
\midrule
w/o free-form reasoning & 24.0 & 76.0 & 50.0 \\
w/o faithfulness reward & 30.0 & 82.0 & 56.0 \\
w/o irrelevant intervention & 32.0 & \textbf{86.0} & 59.0 \\
w/o relevant intervention & 28.0 & 82.0 & 55.0 \\
Training for 1 epoch & 34.0 & 84.0 & 59.0 \\
\midrule
CaM-Wolf (Ours) & \textbf{38.0} & 82.0 & \textbf{60.0} \\
\bottomrule
\end{tabular}
\vspace{-8pt}
\end{table}

Following SLA \citep{10.5555/3692070.3694355} and LSPO \citep{xu2025learning}, we assess our agent's gameplay performance across different team configurations. We examine win rates when the agent represents either Team Village or Team Werewolf in competition. As shown in Table \ref{tab:main_results}, our framework trained on Qwen2.5-72B-Instruct demonstrates a clear advantage when competing against baseline agents utilizing the identical foundation model. Furthermore, when matched against prompt-based baselines powered by the highly capable GPT-4o, CaM-Wolf still achieves competitive and often superior overall results. This highlights the effectiveness of our proposed training method in enhancing the intrinsic reasoning capabilities of the agent.

An observation is that Team Werewolf consistently achieves notably higher win rates compared to Team Village, regardless of the backend LLM or prompting strategy employed. While the informed minority naturally holds a strategic advantage in social deduction games, this extreme performance gap primarily underscores the current limitations of LLM-based agents in handling high-entropy environments. Unlike human players who can employ complex heuristics, these agents often fail to effectively evaluate incomplete, conflicting, and actively deceptive information without a stable ground truth to anchor their reasoning.

Our framework helps mitigate the difficulty of playing with limited information. As shown in Table \ref{tab:main_results}, the largest improvements are observed in the Team Village role. By explicitly modeling the connection between behaviors and underlying logic during training, our agent becomes more effective at recognizing conversational evidence and identifying deceptive patterns compared to baselines.

\begin{table}[t]
\caption{\textbf{Performance against human players.} “Avg. Votes” indicates the average number of votes each agent received per game (maximum 5 votes), and “Avg. Human Votes” represents the average number of votes each agent received from human players per game (maximum 1 vote).}
\vspace{-8pt}
\label{tab:user_study}
\centering
\begin{tabular}{lccc}
\toprule
\textbf{Agent} & \textbf{\makecell{Avg. Votes\\($\downarrow$)}} & \textbf{\makecell{Avg. Human\\Votes  ($\downarrow$)}} & \textbf{\makecell{Win Rate\\($\uparrow$)}} \\
\midrule
Human & 0.31 & --- & 45.2 \\
\midrule
ReAct & 3.45 & 0.88 & 20.0 \\
ReCon & 3.09 & 0.65 & 22.7 \\
LSA & 2.23 & 0.39 & 38.8 \\
LSPO & 0.46 & 0.08 & 42.9 \\
CaM-Wolf & \textbf{0.21} & \textbf{0.05} & \textbf{46.8} \\
\bottomrule
\end{tabular}
\vspace{-8pt}
\end{table}

To further investigate the reasoning capabilities of different agents, we explicitly prompt each agent to predict the roles of all other players after the second discussion round and measure the accuracy of these predictions. As shown in Table~\ref{tab:role_identification}, CaM-Wolf achieves the highest role identification accuracy of 32.6\%, outperforming all baselines. This confirms that our causal-aware training paradigm enhances the agent's ability to infer hidden roles from observable behavioral evidence.

\subsection{Ablation Study}

To validate the effectiveness of individual mechanisms in our framework, we conduct an ablation study. We first examine the impact of the free-form reasoning. When this component is removed, the overall performance experiences a noticeable decline, particularly in the win rate for Team Village. This suggests that allowing the agent to explicitly process information and formulate strategies before making structured role identifications is essential for navigating the complex and deceptive environment of the game.

Furthermore, we evaluate the contribution of our proposed causal reward mechanisms. Eliminating the faithfulness reward entirely or removing either the relevant or irrelevant premise interventions leads to a decrease in the agent's gameplay effectiveness. This highlights the importance of our counterfactual intervention approach, which ensures the model grounds its deductions in genuine behavioral evidence rather than unverified speculation.

Additionally, as outlined in Section \ref{sec:training_scheme}, our model is trained for 2 epochs. We also evaluate the checkpoint saved after a single epoch of training. This variant exhibits a slight drop in performance.

\begin{figure}[t]
  \centering
  \subfigure[]{\includegraphics[width=0.45\linewidth]{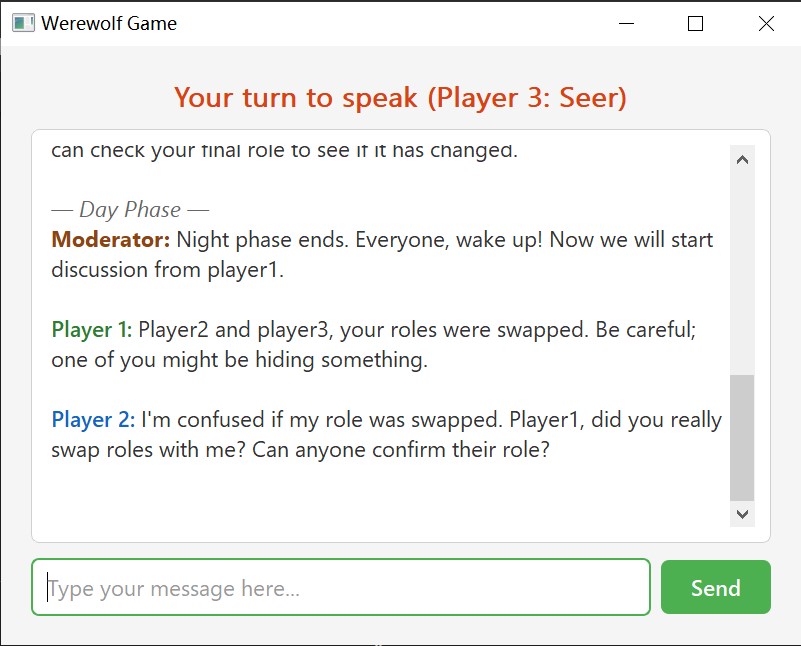}\hspace{0.03\linewidth}}
  \subfigure[]{\includegraphics[width=0.45\linewidth]{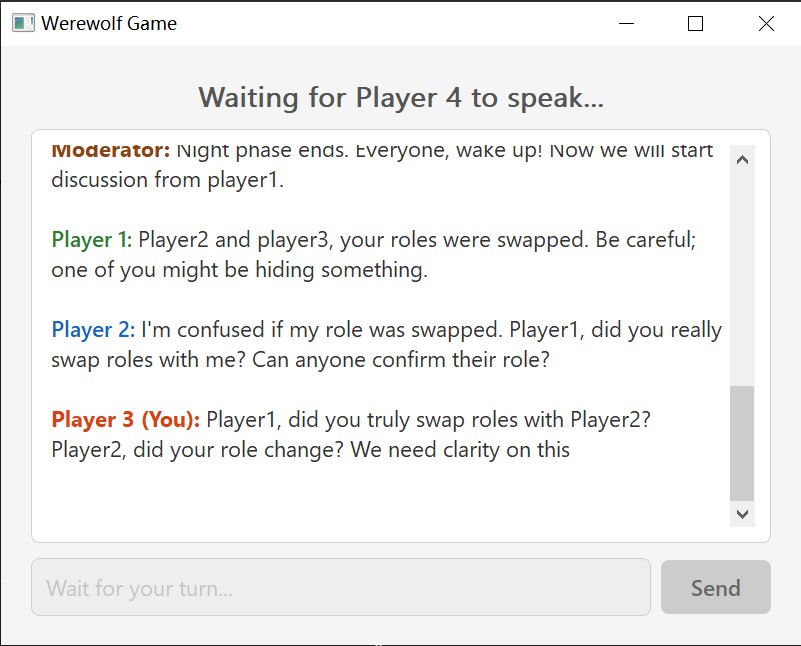}}

\subfigure[]{\includegraphics[width=0.45\linewidth]{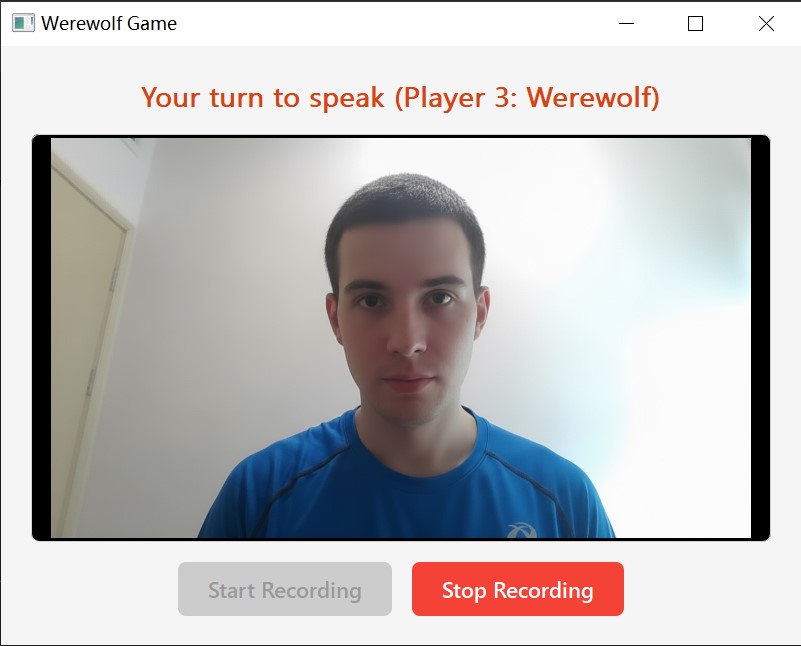}\hspace{0.03\linewidth}}
  \subfigure[]{\includegraphics[width=0.45\linewidth]{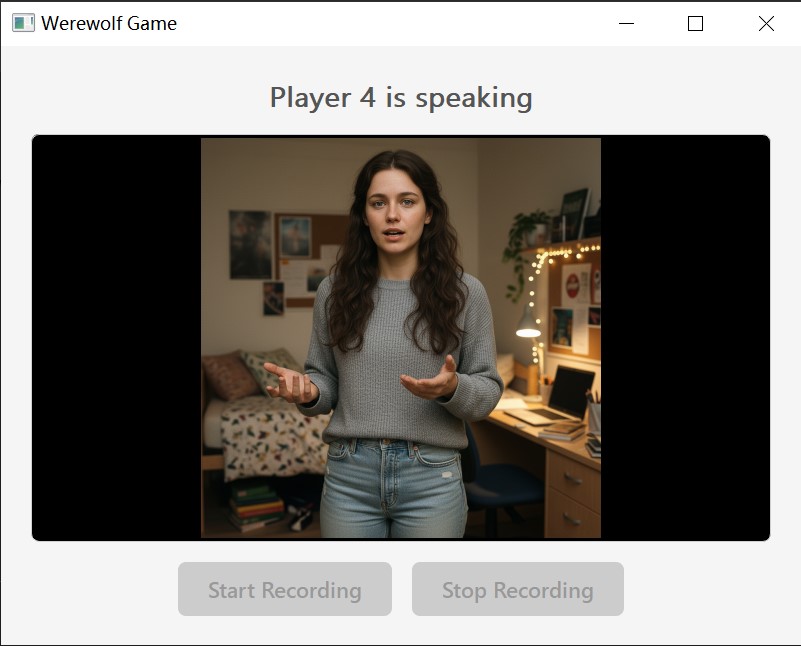}}
  \vspace{-8pt}
  \caption {User interfaces for different interaction paradigms. (a) and (b) show the text-based interaction interface. (c) and (d) illustrate the multimodal interaction interface. The face in (c) has been edited using AI for demonstration purposes only and does not reflect the actual identity or affiliation of any participant or author.}
  \label{fig:gui}
  \vspace{-8pt}
\end{figure}

\subsection{Human Evaluation}
\label{sec:against_human}

To evaluate our agent's performance in realistic human-AI interaction scenarios, we recruit 16 volunteers to participate in our study. Among them, 10 have prior experience playing Werewolf, 4 are familiar with the game rules but have never played, and 2 are completely unfamiliar. Following a comprehensive explanation of the game rules, each participant completes 5 games of 5-player Werewolf. Each game features 1 human player paired with 4 AI agents randomly selected from our agent pool, including CaM-Wolf and four baselines. Detailed game rules are provided in the supplementary materials.

Since the main experiments are conducted in a 7-player setting, we retrain CaM-Wolf using the same method described in Section~\ref{sec:training_scheme} to adapt it to the 5-player configuration. To ensure a fair comparison and enable complete multimodal interaction, we integrated our Perceiver and Performer modules into all baselines.

The results are presented in Table \ref{tab:user_study}. CaM-Wolf achieves the highest win rate among all AI agents. It also receives the fewest votes on average, both overall and from human players specifically, indicating superior suspicion avoidance. This demonstrates our agent's effectiveness in mixed human-AI gameplay scenarios, successfully convincing human players of its trustworthiness while maintaining strong strategic gameplay.

\begin{figure}
    \centering
    \includegraphics[width=\columnwidth]{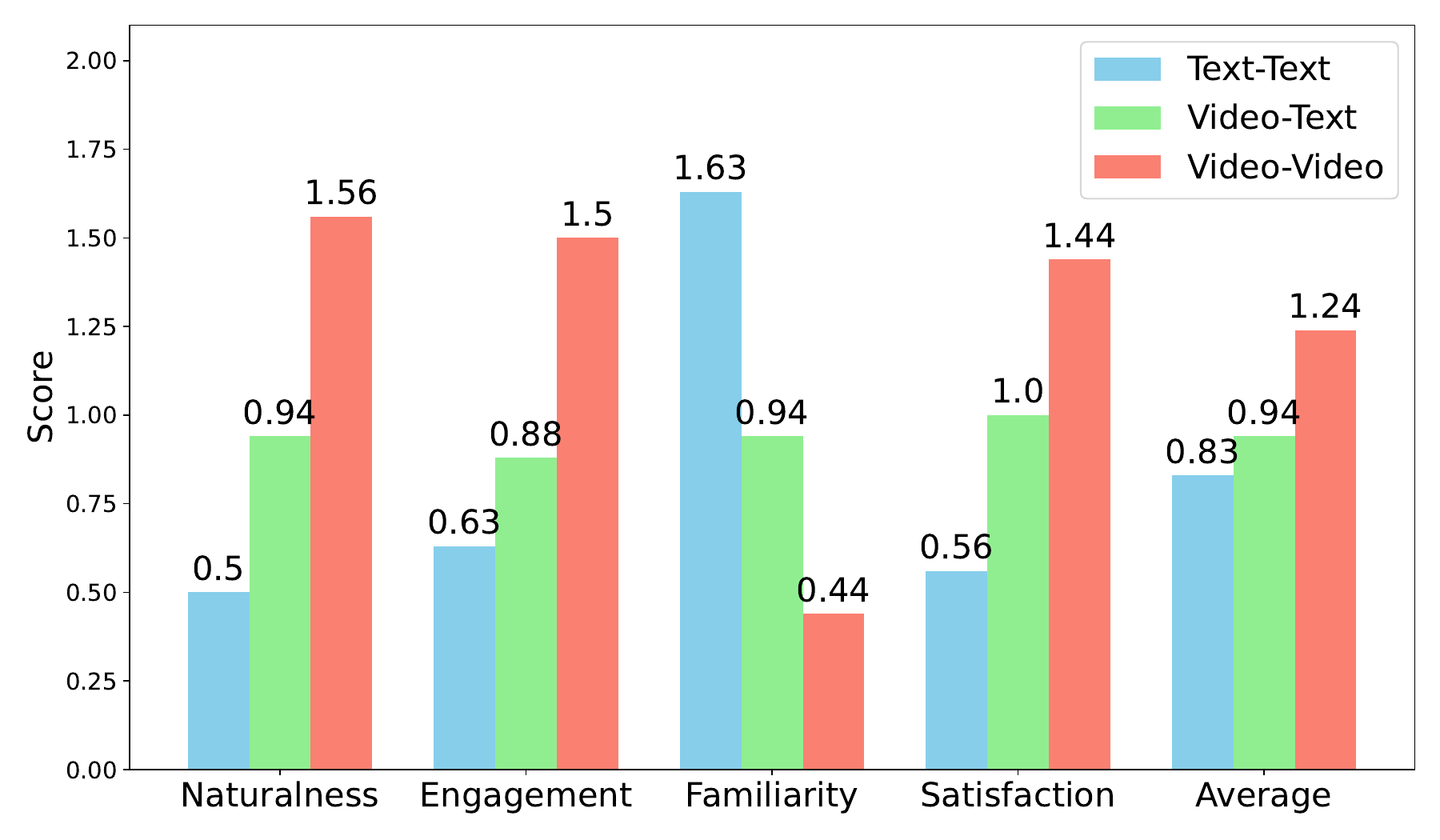}
    \vspace{-8pt}
    \caption{\textbf{Human ranking scores for different interaction paradigms.} Each participant plays 2 games under each of the 3 paradigms and then completes a questionnaire ranking the paradigms on each metric.}
    \label{fig:interaction_mm}
    \vspace{-8pt}
\end{figure}

% \section{User Evaluation}

% To comprehensively assess the user experience and interaction quality of our multimodal system, we recruit an additional 8 volunteers beyond the 8 participants mentioned in Section~\ref{sec:against_human}, bringing our total participant sample to 16 individuals. These additional volunteers consist of 2 females and 6 males, aged between 21 and 28. Among them, 2 have extensive experience with social deduction games, 3 have basic familiarity with game rules but lack practical playing experience, and 3 are completely new to the game. All participants receive comprehensive instruction in game rules and interaction protocols before participating in our studies.

% \subsection{Overall Interaction Experience}

To comprehensively assess the impact of multimodal capabilities on user experience, we evaluate three distinct interaction paradigms that represent different levels of multimodal integration:

\begin{itemize}

\item \textbf{Text-Text.} Participants engage in text-based communication. Human players type their responses using a keyboard and receive information from other players through text displayed on screen.

\item \textbf{Video-Text.} This paradigm introduces multimodal input while maintaining textual output. Participants sit in front of a computer and, during their turn, click buttons to capture video segments of their speech and gestures. These recorded videos are then processed by the agent. However, agent responses are still presented as text.

\item \textbf{Video-Video.} This paradigm implements full bidirectional multimodal communication. Participants provide input through video recording as in the previous paradigm, but agent responses are delivered through animated avatar videos. Participants directly observe and listen to the avatar.
\end{itemize}

Figure~\ref{fig:gui} illustrates the user interfaces for these different interaction paradigms.
We implement a comparative ranking methodology to evaluate user preferences across interaction paradigms. Participants are asked to rank-order the three paradigms on each evaluation dimension from most preferred (first) to least preferred (third). To facilitate quantitative analysis, we apply a weighted scoring system where first-place rankings receive 2 points, second-place rankings receive 1 point, and third-place rankings receive 0 points. The mean scores across all participants are presented in Figure~\ref{fig:interaction_mm}.

% \begin{figure*}
%     \centering
%     \includegraphics[width=\textwidth]{figures/2dto3d.pdf}
%     \caption{Examples of 3D avatar models generated from 2D images. The top row shows the original 2D avatar images, while the bottom row displays the corresponding 3D models generated.}
%     \label{fig:3d}
% \end{figure*}

The Video-Video paradigm achieves the highest average score, significantly outperforming both Video-Text and Text-Text. This superiority is particularly pronounced in naturalness, engagement, and satisfaction, suggesting that bidirectional multimodal communication substantially enhances the user experience by providing more intuitive and immersive interactions.

\section{Conclusion}

In this paper, we introduced CaM-Wolf, the first SDG agent that integrates multimodal perception and generation capabilities. CaM-Wolf processes video inputs from human players and presents itself through an animated avatar, creating more natural and immersive human-AI interactions. Through a causal-aware Reasoner trained via reinforcement learning, CaM-Wolf intrinsically generates faithful and evidence-grounded reasoning by identifying causal relationships between players' observable behaviors and their hidden roles. Our experiments and user study demonstrates that CaM-Wolf improves gameplay performance and enhances the quality of human-AI interaction.

\begin{acks}
This work is supported by the National Natural Science Foundation of China (No. 62406267), Guangdong Provincial Project (No. 2024QN11X072), Guangzhou Municipal Science and Technology Project (No. 2025A04J4070) and CAAI-Tencent Rhino-Bird Open Research Fund.
\end{acks}

\bibliographystyle{ACM-Reference-Format}
\balance
\bibliography{sample-base}

\end{document}